%% file: main.tex
\documentclass[transmag]{IEEEtran}

\usepackage{algorithm}
\usepackage{multirow}
\usepackage{makecell}
\usepackage{amsfonts}
\usepackage{booktabs}
\usepackage{color}

\usepackage{subfigure}

\usepackage{times}
\usepackage{epsfig}
\usepackage[cmex10]{amsmath}
\usepackage{amssymb}
\usepackage{multirow}
\usepackage{stfloats}
\usepackage{tabularx}
\usepackage{pifont}
\usepackage{makecell}
\usepackage{threeparttable}
\usepackage{mathtools}
\usepackage{xcolor}
\usepackage{graphicx}
\usepackage{caption}
\usepackage{algorithmic}

\usepackage{url}

\begin{document}
	%
	\title{Weakly-Supervised Multi-Face 3D Reconstruction}
	
	\author{
		Jialiang~Zhang, Lixiang~Lin, Jianke~Zhu~\IEEEmembership{Senior~Member,~IEEE}, Steven~C.H.~Hoi~\IEEEmembership{Fellow,~IEEE}
		\IEEEcompsocitemizethanks{
			\IEEEcompsocthanksitem Jialiang~Zhang, Lixiang~Lin, and Jianke~Zhu are with the College of Computer Science and Technology, Zhejiang University, Hangzhou, China, 310027.  Jianke Zhu is also with the Alibaba-Zhejiang University Joint Research Institute of Frontier Technologies, Hangzhou, China.\protect\\ 
			E-mail: \{zjialiang, lxlin, jkzhu\}@zju.edu.cn.
			\IEEEcompsocthanksitem Steven C.H. Hoi is with Salesforce Research, Singapore.\protect\\
			E-mail: chhoi@smu.edu.sg
			\IEEEcompsocthanksitem Jianke Zhu is the Corresponding Author.
		}
		\thanks{}}
	\maketitle
	
\begin{abstract}
3D face reconstruction plays a very important role in many real-world multimedia applications, including digital entertainment, social media, affection analysis, and person identification. The de-facto pipeline for estimating the parametric face model from an image requires to firstly detect the facial regions with landmarks, and then crop each face to feed the deep learning-based regressor. Comparing to the conventional methods performing forward inference for each detected instance independently, we suggest an effective end-to-end framework for multi-face 3D reconstruction, which is able to predict the model parameters of multiple instances simultaneously using single network inference. Our proposed approach not only greatly reduces the computational redundancy in feature extraction but also makes the deployment procedure much easier using the single network model. More importantly, we employ the same global camera model for the reconstructed faces in each image, which makes it possible to recover the relative head positions and orientations in the 3D scene. We have conducted extensive experiments to evaluate our proposed approach on the sparse and dense face alignment tasks. The experimental results indicate that our proposed approach is very promising on face alignment tasks without fully-supervision and pre-processing like detection and crop. Our implementation is publicly available at \url{https://github.com/kalyo-zjl/WM3DR}.
	
\end{abstract}
	
\begin{IEEEkeywords}
	Multi-Face Reconstruction, Weakly-Supervised Learning, Single-Shot
\end{IEEEkeywords}
	
	%
\IEEEpeerreviewmaketitle

\section{Introduction}\label{sec:introduction}

3D face modeling plays a very important role in many real-world multimedia applications, including digital entertainment~\cite{thies2016face2face}, social media~\cite{zhou2019talking},  affection analysis and person identification~\cite{zulqarnain2018learning}.

With the prevalence of deep learning, 3D face reconstruction from a single image becomes promising, in which the deep models are learned without 3D labels in the unsupervised or weakly-supervised fashion~\cite{genova2018unsupervised,tewari2018self,tran2018nonlinear,deng2019accurate, wu2020unsupervised}. Nevertheless, these methods assume that the face has already been rectified into the predefined frame. The de-facto pipeline is made of two consecutive steps. Firstly, it locates the facial regions and landmarks using the off-the-shelf object detector~\cite{xiang2017joint}. Then, each face is cropped to feed the deep learning-based regressor~\cite{zhu2016face}, independently. The whole process requires one neural network forward for the detection task and $n$ inference passes to predict the 3D face model parameters, where $n$ is the total number of faces found within an image. Obviously, there are redundant computations on extracting deep learning features to estimate the model coefficients.  

In contrast to considerable research efforts on single face reconstruction~\cite{bas20173d,dou2017end,genova2018unsupervised,tewari2018self,tran2018nonlinear}, there is few work that jointly recovers multiple faces from an image. Bindita \emph{et al.}~\cite{chaudhuri2019joint} predict the bounding boxes and decouple the pose, identity, and expression parameters for multiple faces. Besides the detection and sparse landmark regression modules, Deng \emph{et al.}~\cite{deng2020retinaface} directly regress the dense 3D face points on the image plane with an extra branch. Although having achieved promising results, they only recover the face shape while ignoring the texture and lighting information. More importantly, these approaches are trained in the supervised manner that requires the ground-truth geometry as the regression targets. As the training samples are either synthesized by rendering or be obtained from the conventional optimization-based method~\cite{blanz1999morphable}, they are usually less robust to large variations on pose and lighting.

\begin{figure}[t]
	\begin{center}
		\includegraphics[width=0.5\textwidth]{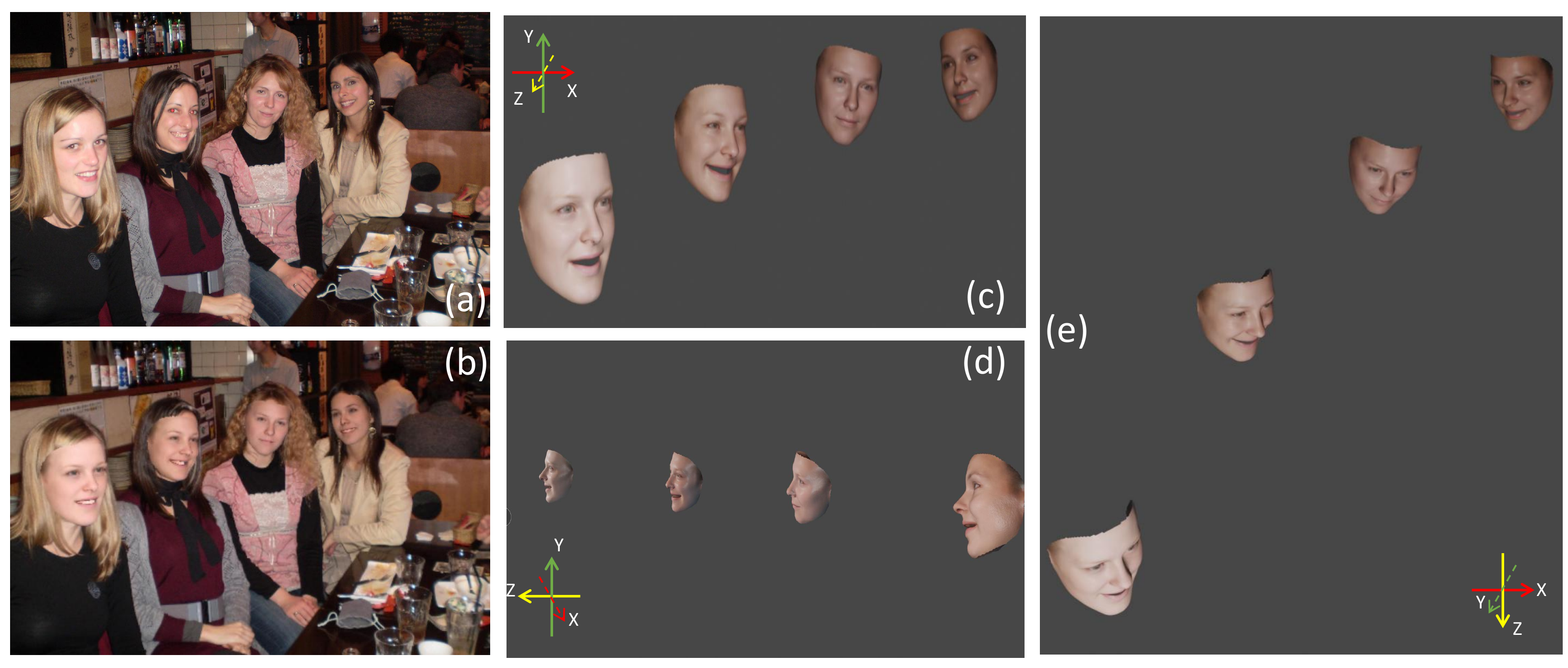}
	\end{center}
	\vspace{-0.1in}
	\caption{Multi-face reconstruction from different perspectives. (a) The input image. (b) The reconstructed faces. (c) Frontal view. (d) Right view. (e) Top view.}	\label{fig:ill}
\end{figure}

Generally, multiple faces within an image should share the same intrinsic camera parameters, as shown in Fig.~\ref{fig:ill}. This means that the focal length and camera center coordinates in their recovered 3D parametric models are kept the same. However, most of the conventional detection and reconstruction pipeline~\cite{chaudhuri2019joint,deng2020retinaface} estimates the camera model for each face in the image, independently. Therefore, they ignore the global head positions and orientations for each individual in the 3D scene, which are the high-level semantic information for image understanding.

To address the above limitations, in this paper, we propose a single-shot multi-face reconstruction framework in a fully weakly-supervised fashion. Comparing to the conventional methods performing forward inference for each detected instance separately, we suggest an effective end-to-end framework for multi-face 3D reconstruction, which is able to predict the model parameters for multiple instances using single network inference. Therefore, our proposed approach not only greatly reduces the computation redundancy in the conventional pipeline but also makes the deployment procedure much easier. More importantly, we employ the same global camera model for the reconstructed faces in each image, which makes it possible to recover the relative head positions and orientations in the 3D scene, as shown in Fig.~\ref{fig:ill}. To this end, a hybrid loss upon each facial feature point is employed to learn the decoupled 3D face model parameters in the weakly-supervised manner, which is constrained by the sparse 2D facial landmarks and original image with a differentiable image formation procedure~\cite{ravi2020pytorch3d}. Finally, we have conducted the extensive experiments to evaluate our proposed approach on both sparse and dense face alignment tasks. The experimental results indicate that our proposed approach is very promising on face alignment without fully-supervision and pre-processing like detection and crop.

From above all, the main contributions of this paper can be summarized as follows: 1) an efficient weakly-supervised single-shot multi-face reconstruction framework that can estimate the model coefficients of multiple instances using single forward pass; 2) a unified camera model that is able to recover the relative head positions and orientations in 3D scene; 3) a stage-wise training scheme to optimize the overall framework which trains a robust face model in the end-to-end fashion. Extensive experiments including sparse and dense face alignment as well as the qualitative results demonstrate the efficacy of our proposed framework.

The rest of this paper is organized as follows. Section II reviews the related work. Section III presents our proposed weakly-supervised single shot multi-face reconstruction framework. Section IV gives our experimental evaluation in detail and
finally Section V concludes this work.

\section{Related Work}

During the past decade, extensive research efforts have been devoted to 3D face reconstruction, which aims at recovering the 3D facial geometry from image collection or even with single image. In general, these approaches can be roughly categorized into two groups. One is the model-free methods that are based on multiview geometry or stereo rigs ~\cite{fyffe2014driving, wu2020unsupervised}. The other relies on the 3D deformable model~\cite{blanz1999morphable, paysan20093d}, which formulates 3D face reconstruction as the problem of parameter estimation.

In this paper, we focus our attention on the model-based approaches, which have already achieved encouraging progress. Blanz and Vetter~\cite{blanz1999morphable} build the first 3D Morphable Models (3DMM) from the laser scans, which is the most widely used face model. It employs the linear subspace to represent the shape and texture through principal component analysis (PCA). To deal with expression variations, Zhou \emph{et al.}~\cite{cao2013facewarehouse} extend the PCA space with the manually created blending shapes. Based on the deformable models, they directly estimate the 3DMM coefficients and image formation parameters through nonlinear least squared optimization~\cite{blanz1999morphable}. 

In contrast to the conventional model-fitting approaches, deep learning-based methods~\cite{bas20173d,dou2017end,genova2018unsupervised,richardson20163d,tran2017regressing,tewari2017mofa} have recently received quite a bit of attention. The supervised methods directly regress PCA coefficients~\cite{bas20173d,tran2017regressing,richardson20163d} using the ground-truth of paired 2D-3D data that is usually synthesized by rendering~\cite{richardson20163d}. Therefore, they may not perform well in dealing with illumination variations and occlusions in the wild. Building the ground-truth data through the iterative optimization~\cite{tran2017regressing} still leads to the domain gap, where the learned face model is delicate in the unconstrained scenarios. 

To circumvent the above issue, the weakly-supervised and unsupervised approaches have been proposed~\cite{genova2018unsupervised,tewari2018self,tran2018nonlinear,deng2019accurate}, which greatly alleviate the problem of lacking $3$D supervision for single face reconstruction. The discrepancy between the input image and the rendered counterpart is considered as supervision, which relies on a differentiable image formation procedure~\cite{Genova_2018_CVPR,henderson19ijcv}. Tewari \emph{et al.}~\cite{tewari2017mofa} and Sengupta \emph{et al.}~\cite{sengupta2018sfsnet} treat pixel-wise photometric difference as measurement. Additional face image discrepancy on the perception-level is adopted in training loss~\cite{Genova_2018_CVPR,deng2019accurate}, which is measured by the Euclidean distance between deep features extracted from the face recognition network. To improve the robustness during the training procedure, the sparse landmark loss is usually adopted as an auxiliary constraint. Although having achieved encouraging progress, all these methods assume that the facial region has already been detected and cropped. 

To recover the multiple face models in image, the conventional pipeline~\cite{bas20173d,dou2017end,genova2018unsupervised,richardson20163d,tran2017regressing,tewari2017mofa} requires to detect the faces firstly, and then reconstruct each face model, separately. Such scheme employs an extra pre-processing step, where the computational time increases linearly with the total number of located faces. The separated steps not only make the whole pipeline redundant but also involve with several different components for deployment. To this end, some approaches are proposed to reconstruct multiple faces simultaneously, which enjoys the merit of sharing feature representation. Moreover, it reduces the computational cost and improves performance for each task. Bindita \emph{et al.}~\cite{chaudhuri2019joint} propose to jointly predict the bounding box and recover $3$D shape for each face, which employ $3$DMM with the attributes like identity and expression. Deng \emph{et al.}~\cite{deng2020retinaface} directly regress the dense 3D facial points on the image plane for each individual. However, the pose information cannot be predicted by the network inference, the further least-squares optimization should be applied for each face to estimate the pose. Although having achieved the promising results, these methods focus on recovering the $3$D facial geometry while ignoring the texture that is essential to a large amount of real-world applications. Additionally, they are dependent on $3$D supervision during training, which makes them inconvenient to handle the facial images in-the-wild.

\input{soft}
\input{experiment}
\input{conclusion}

\bibliographystyle{abbrv}
\bibliography{ref}

\ifCLASSOPTIONcaptionsoff
\newpage
\fi

\end{document}

%% file: soft.tex
\begin{figure*}[htbp]
	\begin{center}
		\includegraphics[width=.8\textwidth]{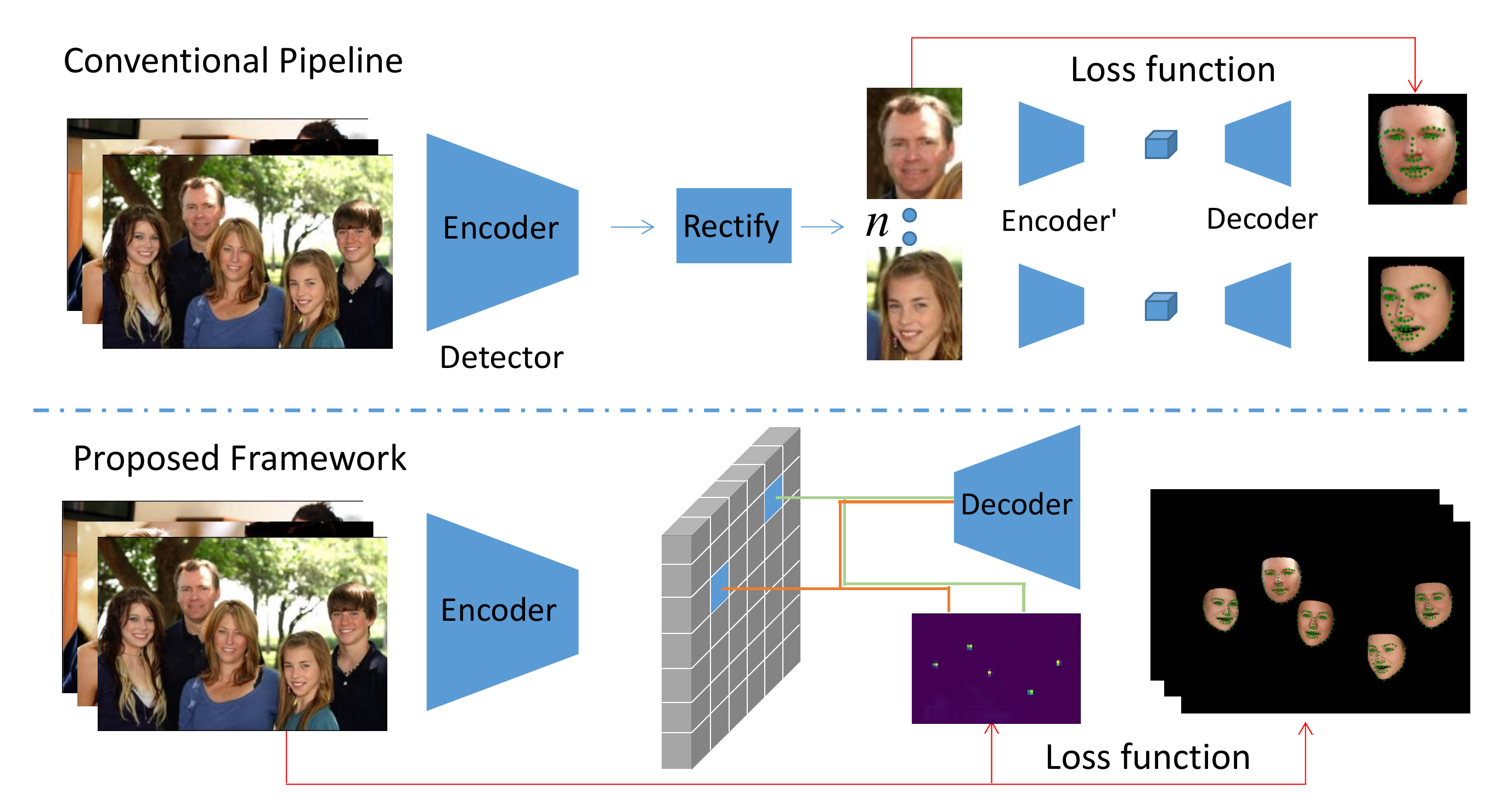}
	\end{center}
	\caption{Comparing our proposed framework against the conventional pipeline. The top row shows that the conventional pipeline needs firstly to detect and align the faces, and then reconstructs each item, separately. The bottom row illustrates our proposed weakly-supervised single-shot multiple-face reconstruction framework, which takes advantage of an end-to-end auto-encoder network structure.}
	\label{fig:architecture}
\end{figure*}
\section{Weakly-Supervised Multi-Face 3D Reconstruction}

In this section, we first briefly overview our proposed weakly-supervised multi-face 3D reconstruction framework. Then, we present the objective function of the proposed neural network. Finally, we give the detailed implementation on training and inference.

\subsection{Overview}

Generally, the model-based 3D face reconstruction framework shares an encoder-decoder structure. ResNet-50~\cite{he2016deep} is employed as the encoder, which is one of the most popular neural network backbones. Moreover, $3$D Morphable Model ($3$DMM) is treated as the decoder to reconstruct the facial shape and texture. 

As illustrated in Fig.~\ref{fig:architecture}, the conventional pipeline sequentially performs detection and single face reconstruction, whose inference time $T_c$ is composed of two separate parts as follows:
\begin{equation}
	T_{c}=T_{Encoder} + n\cdot(T_{Rectify} + T_{Encoder'}+T_{Decoder})
\end{equation}
where $n$ represents the total number individuals located in image. ``Encoder" is essentially a face detector. $T_{Encoder}$ denotes the computational time on detection, which is made of single network forward of auto-encoder along with non-maximal suppression. $T_{Rectify}$ is made of the computation cost spent on cropping and resizing each face region to feed the deep regressor.

As for our proposed weakly-supervised single-shot multi-face reconstruction framework, we aim at jointly recovering multiple faces using a single forward pass. By taking advantage of an end-to-end auto-encoder network structure, it recovers the parameters of multiple instances through the shared network ``Encoder". It predicates both the face locations and their corresponding $3$DMM coefficients simultaneously. Thus, the inference time of our proposed approach can be derived as below:
\begin{equation}
	T=T_{Encoder} + n\cdot T_{Decoder}
\end{equation}

Assume that the face detector employs the same backbone as ``Encoder'", the conventional pipeline requires $n+1$ inference of ResNet-50 while our proposed framework only has single forward pass. Therefore, our presented method is quite efficient in the case of multiple individuals in image.

\subsubsection{Encoder}
As ResNet-50 is adopted as the backbone neural network, the final feature map $\Phi$ of shared network ``Encoder" has a stride of $8$ through feature aggregation.

\begin{figure}[htbp]
	\begin{center}
		\includegraphics[width=.5\textwidth]{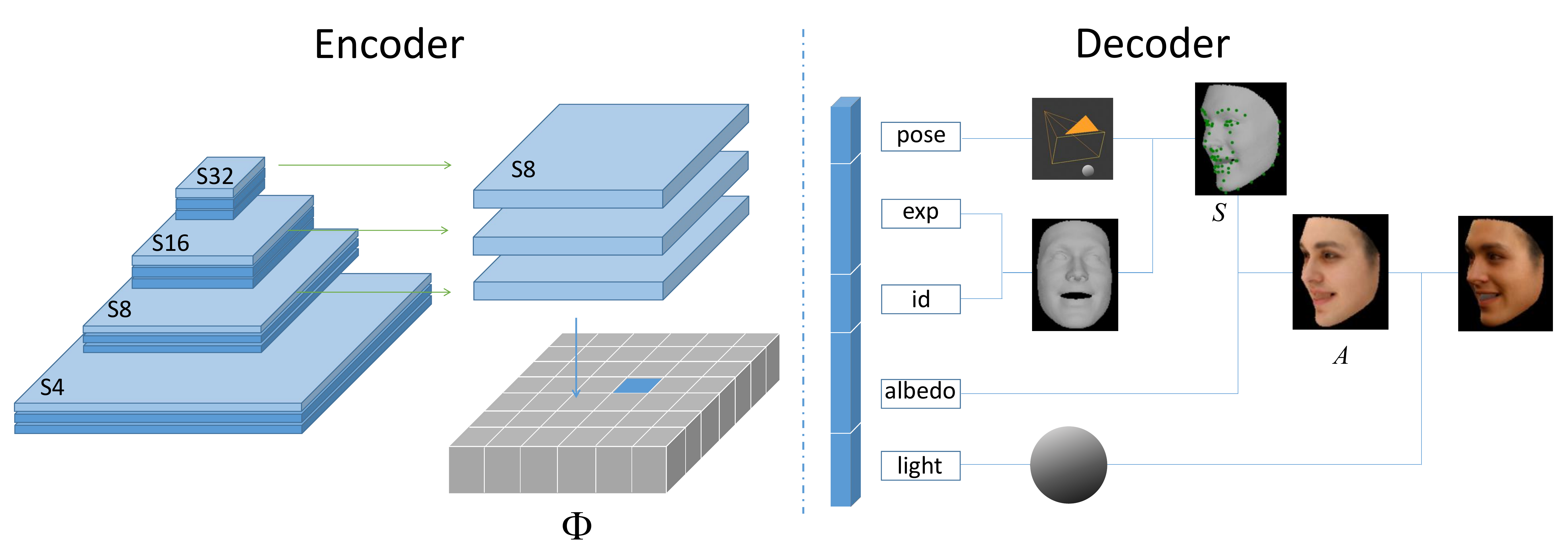}
	\end{center}
	\caption{The details of network Encoder and Decoder.}
	\label{fig:EDecoder}
\end{figure}

As shown in Fig.~\ref{fig:EDecoder}, each point in feature map $\Phi$ is a vector containing all the necessary information to determine the details about one face, which is further decoded in order to render the $3$D reconstruction results. In addition to the conventional $3$DMM coefficients, we estimate a vector $\delta_{m} \in \mathbb{R}^{1}$ to determine the location of each face. In summary, we denote all the parameters to be encoded as $\delta = (\delta_{m}, \delta_{id}, \delta_{exp}, \delta_{alb}, \delta_{pose}, \delta_{illum}) \in \mathbb{R}^{258}$, which are described in the following.

\subsubsection{3D Morphable Model as Decoder} \label{section: face reconstruction}
$3$D Morphable Model ($3$DMM) provides a parametric method to synthesize faces that are represented by their shape and albedo:
\begin{equation}
\begin{aligned}
S &= \bar S + \delta_{id} * S_{id} + \delta_{exp} * S_{exp} \\
A &= \bar A + \delta_{alb} * A_{alb}
\end{aligned}
\end{equation}
where $S \in \mathbb{R}^{3 \times N}$ is the $3$D face shape with $N$ vertices. As in~\cite{deng2019accurate}, $N$ is 35,709. $\bar S \in \mathbb{R}^{3 \times N}$ is the mean neutral shape, $\delta_{id}$ and $\delta_{exp}$ are the parametric coefficients corresponding to the identity basis $S_{id}$ and expression basis $S_{exp}$, respectively. Similarly, the facial albedo $A \in \mathbb{R}^{3 \times N}$ contains the colors of $N$ vertices for shape $S$. $\bar A$ is the mean albedo, and $\delta_{alb}$ is the texture coefficients with respect to the albedo basis $A_{alb}$. We adopt a subset of these basis from~\cite{blanz1999morphable} and~\cite{cao2013facewarehouse}, where $\delta_{id} \in \mathbb{R}^{80}$, $\delta_{exp} \in \mathbb{R}^{64}$, and $\delta_{alb} \in \mathbb{R}^{80}$. To deal with the lighting changes, a vector $\delta_{illum} \in \mathbb{R}^{27}$ is estimated to approximate the scene illumination with Spherical Harmonics(SH).

\subsubsection{Camera Model}

The $3$DMM coefficients are employed to reconstruct the $3$D face model in the canonical shape with albedo. Moreover, the additional information like the camera and head pose is required to render the photo-realistic facial images using the recovered 3D model parameters. 

Instead of dealing with the view parameters of each face separately like the conventional pipeline, a global camera model of perspective projection is employed to project the $3$D face model onto the $2$D image plane. Assume that each face has a similar size in 3D space, and the scale variations in the image plane are mainly due to the foreshortening effects of perspective projection. The intrinsic matrix $K$ is denoted as follows:
\begin{equation}
	K = 
	\left[ \begin{matrix}
		f_g & 0 & w/2 \\
		0 & f_g & h/2 \\ 
		0 & 0 & 1 \\
	\end{matrix}\right]
\end{equation}
where $w$, $h$ are the width and height of the image. $f_g$ is the focal length, which is kept constant for all images. The camera position is determined by the rotation matrix $R \in \mathbb{R}^{3\times3}$ and the translation vector $t \in \mathbb{R}^{3}$, in which $R$ is reconstructed from the predicted rotation vector $\delta_{rot} \in SO(3)$, and $t = [t_x, t_y, t_z]^T$ is explicitly decoded by the predicted translation vector $\delta_{trans} = [d_x, d_y, d_z]^T$ as follows:
\begin{equation}
	\left[ \begin{matrix}
		t_{x} \\
		t_{y} \\ 
		t_{z} \\
	\end{matrix}\right]  = 
	\left[ \begin{matrix}
		d_z(d_x+c_x-w/2)/f_g \\
		d_z(d_y+c_y-h/2)/f_g \\ 
		d_z \\
	\end{matrix}\right]
\end{equation}
where $(c_x, c_y)$ is the center of each face in the image plane. The scale and displacement of each face are implicitly encoded in $t$. Then, the projection between the $3$D face point $X=(x, y, z)^T \in S$ and its corresponding $2$D point $p = (u, v, 1)^T$ in image space is as below:
\begin{equation}
	p \propto K(RX+t)
\end{equation}
The pose parameters form a vector $\delta_{pose} = (\delta_{rot}, \delta_{trans})$, which is predicted by the network inference.

Note that the global perspective projection model enjoys the following advantages. Firstly, the image should have a single unified camera model, e.g the intrinsic matrix $K$, rather than each face region has a local projection model. Secondly, it is easy to incorporate the constant intrinsic matrix into our proposed framework. Finally, the relative positions of reconstructed faces can be retained in the camera space. 

\subsection{Objective Function} \label{section: loss function}
In the following, we introduce the objective function to be optimized in our proposed framework.

Estimating $\delta_{m}$ essentially is a binary classification task. Specifically, a point at the center of the face region should be categorized into the class ``face'', and the remaining points should be labeled as ``background''. Thus, the center loss function is formulated as pixel-wise logistic regression with focal loss via the supervision of center location:
\begin{equation}
	L_{c} = -\frac{1}{n}\sum_{i=1}^{W/r}\sum_{j=1}^{H/r}(1-\hat{\delta}_{m}^{ij})^{\gamma}log(\hat{\delta}_{m}^{ij}),\\
	\label{eq:clsloss}
\end{equation}
where $n$ is the total number of faces in image. Let $\delta_{m}^{ij}\in[0,1]$ denote the predicted probability indicating whether the location $(i,j)$ is a positive instance or not.
$\hat{\delta}_{m}^{ij}$ is defined as follows:
\begin{equation}
	\hat{\delta}_{m}^{ij}=
	\begin{cases}
		{\delta}_{m}^{ij}  & \mbox{if $y^{ij}=1$}\\
		1-{\delta}_{m}^{ij}& \mbox{otherwise,}
	\end{cases}
\end{equation}
where $y^{ij}=1$ is the ground-truth label representing the positive location, and $0$ for the negative one. The hyper-parameters $\gamma$ is set to $2$.

Besides face location $\delta_{m}$, the ground-truth of remaining parameters is not available for images in the wild. To this end, we employ a hybrid loss to train the proposed neural network in a weakly-supervised manner, which includes the pixel-wise loss, perception-level loss, sparse landmark reprojection loss, and regularization loss.

The pixel-wise loss aims at minimizing the discrepancy between the original image and its rendered counterpart. For in-the-wild images, there exist some challenges, such as glasses, make-up, etc, which may lead to the deteriorated performance. To tackle this critical issue, we only take consideration of the skin area during training, which is predicted by the face parsing network\footnote{https://github.com/zllrunning/face-parsing.PyTorch.git The code to generate the face parsing result}. Specifically, the pixel-wise loss adopts a robust $l_{2,1}$ loss as follows:
\begin{equation}
	L_{pix} = \frac{\sum_{i=1}^{W/r}\sum_{j=1}^{H/r} M(i,j) \left\|I(i,j)-I^*(i,j)\right\|_2}{\sum_{i=1}^{W/r}\sum_{j=1}^{H/r} M(i,j)}, 
\end{equation}
\begin{equation}
	M=M_{skin} \circ M_{mask}
\end{equation}
where $I$ and $I^*$ are the rendered and original images, respectively. $\circ$ denotes Hadamard product. $M_{skin}$ and $M_{mask}$ are the union set of skin and mask area of $n$ faces in the image. The distance in color space is based on $l_{2}$-norm and the summation over all pixels enforces the sparsity constraints based on $l_1$-norm.

In contrast to the pixel-wise loss that is enforced in the RGB color level, we employ the perception-level loss to minimize the Euclidean distance between high-level deep features extracted from a face recognition network. Specifically, the perception-level loss is defined as below:
\begin{equation}
	L_{per} = \frac{1}{n} \sum_{k=1}^{n} \left\| \phi(I_{k}) - \phi(I^*_{k}) \right\|^2_2,
\end{equation}
where $I_{k}, I^*_{k}$ are the $k$-th face region of the image $I, I^*$. $\phi(\cdot)$ is a deep feature extractor~\cite{schroff2015facenet}.

To enable the model with better generalization capability on head pose and face shape, we employ the sparse landmark reprojection error measured by Euclidean distance as follows:
\begin{equation}
	L_{lan} = \frac{1}{n} \sum_{k=1}^{n} \sum_{j=1}^{m} \omega_{j} \left\| q_{k,j} - q_{k,j}^* \right\|_2^2,
\end{equation}
where $m=68$ is the total number of landmarks for each face. $q_{k,j}$ is the $j$-th projected landmark of the $k$-th face. $q_{k,j}^*$ is the corresponding ground truth. $\omega_{j}$ is the weight for the $j$-th landmark, which is $20$ for mouth and nose points and $1$ for others. 

In addition to the aforementioned losses that are weakly-supervised to reconstruct $3$D faces from image, it is necessary to enforce a few regularization terms to prevent the model degeneration:
\begin{equation}
	L_{norm} = \frac{1}{n} \sum_{k=1}^{n} (\lambda_{id} \left\| \delta_{id,k} \right\|^2_2 + \lambda_{exp} \left\| \delta_{exp,k} \right\|^2_2 + \lambda_{alb} \left\| \delta_{alb,k} \right\|^2_2),
\end{equation}
where $L_{norm}$ enforces a mean face prior distribution. $\lambda_{id}, \lambda_{exp}, \lambda_{alb}$ are set to $1.0, 0.8, 0.0017$, as in~\cite{deng2019accurate}.

To favor a constant skin albedo, we enforce a flattening constraint to penalize the texture map variance similar to~\cite{deng2019accurate}. Let $\rm Var$ denote the variance function. The texture variance loss can be derived as below:
\begin{equation}
	L_{var} = \frac{1}{n} \sum_{k=1}^{n} \sum_{c \in \{r,g,b\}} {\rm Var}(A_{k,c,\mathcal{R}})
\end{equation}
where $\mathcal{R}$ is the pre-defined subset of albedo $A$ covering skin region including cheek, nose, and forehead. 

Then, the regularization term is a weighted sum of $L_{norm}$ and $L_{var}$, which is denoted as below:
\begin{equation}
	L_{reg} = \lambda_{norm}L_{norm} + \lambda_{var}L_{var},
\end{equation}
where $\lambda_{norm}$ is $0.0001$, and $\lambda_{var}$ is set to $0.001$.

In summary, the overall objective function can be denoted as follows:
\begin{equation}
	L = \lambda_{c}L_{c} + \lambda_{pix}L_{pix} + \lambda_{per}L_{per} + \lambda_{lan}L_{lan} + L_{reg},
\end{equation}
where $\lambda_{c}, \lambda_{pix}, \lambda_{per}, \lambda_{lan}$ are set to $1, 100, 0.01, 0.1$, respectively.

\subsection{Implementation}
In the following, we describe the detailed implementation for training and inference.

{\bf Training.} It is not trivial to directly optimize the whole framework with $L$ due to involving with the optimization in different feature spaces, in which directly optimizing $L$ leads to poor convergence in the training process. 

To tackle this problem, we propose a stage-wise optimization scheme to better build a robust multi-face reconstruction model. At the first stage, we focus on the face reconstruction. Specifically, $\lambda_c$ is set to $0$ in $L$, and the input is set to $224\times224$ with single face aligned at the center of the image. After achieving the convergence at the first stage, we empower the network with the capability of localization and reconstruction for various scales. At the second stage, we enable the center loss $L_c$ by setting $\lambda_c$ is set to $1$. The training input is the augmented multi-face image with input size $512\times512$, accordingly. Then, the network is optimized with full loss $L$ to exploit the capability of joint reconstruction for multiple faces.

{\bf Inference.} The trained model takes an input image with the size of a multiply of $32$. Through the single forward pass, it is able to reconstruct all faces in the image. Our model does not explicitly estimate the bounding box for each face. As in~\cite{zhou2019objects}, we simply extract the local peaks of $\delta_m$ in the corresponding map without any post-process step like NMS, which makes the whole framework very concise.

%% file: experiment.tex
\section{Experiments} 
In this section, we firstly describe the details of the experimental setup. Then, we conduct experiments on single face reconstruction, where both sparse and dense face alignment are examined. Finally, we evaluate our proposed framework on multi-face images and demonstrate the qualitative results of reconstructed $3$D faces.

\subsection{Experimental Setup}

\textbf{Dataset.} We collect a corps of face images from multiple sources, including CelebA~\cite{liu2015deep}, 300W LP~\cite{zhu2016face}, and  300VW~\cite{shen2015first}.
For those datasets without ground truth annotations, we employ the off-the-shelf facial keypoint detector~\cite{bulat2017far} to locate the sparse landmarks. Most of these datasets contain only a single face. Although some dataset like WIDERFACE~\cite{yang2016wider} has multiple faces in an image, which is mainly built for face detection rather than 3D reconstruction. Since it contains a large number of tiny blurring faces, which is hard to use in practice. To tackle the issue of lacking the training data with multiple faces, we randomly augment the single face images to synthesize a larger sample, which may have $1-10$ faces with various scales randomly placed in the image frame without highly overlapping among each other.

We take ResNet-50 as the backbone, which is pre-trained from ImageNet.
The whole neural network is optimized using Adam~\cite{kingma2014adam}. The first stage is trained with the batch size of $128$ paralleled on $4$ RTX$2080$Ti for $200$K iterations with the initial learning rate 0.0001, decayed by $10$ at $150$K-th iteration. The second stage is finetuned from stage one with the batch size of $32$ paralleled on $4$ RTX$2080$Ti for $100$K iterations with the initial learning rate 0.0001, decayed by $10$ at the $50$K-th iteration.

\subsection{Results on Single Face Reconstruction}
To demonstrate the effectiveness of single face reconstruction, we evaluate our proposed approach on the AFLW2000-3D dataset\cite{zhu2016face}, which is built to examine 3D face alignment performance.  It contains the ground truth 3D faces and their corresponding 68 landmarks. We evaluate the Normalized Mean Errors(NME) on both sparse and dense points.

\begin{table}[!ht]
	\small
	\centering
	\caption{Comparison of NME (\%) of 68 landmarks on AFLW2000-3D with respect to three groups based on yaw angles.}
	\resizebox{.45\textwidth}{!} {
	\begin{tabular}{c||c|c|c|c}
		\hline
		\multirow{2}{*}{\textbf{Method}} & \multicolumn{4}{c}{\textbf{AFLW2000-3D (68 pts)}} \\ \cline{2-5} 
		& {[}0, 30{]} & {[}30, 60{]} & {[}60, 90{]} & Mean \\ \hline
		ESR~\cite{cao2014face} & 4.60 & 6.70 & 12.67 & 7.99 \\ 
		SDM~\cite{xiong2015global} & 3.67 & 4.94 & 9.67 & 6.12 \\ 
		3DDFA~\cite{zhu2016face} & 3.78 & 4.54 & 7.93 & 5.42 \\ 
		Yu \emph{et al.}~\cite{yu2017learning} & 3.62 & 6.06 & 9.56 & 6.41 \\ 
		DeFA~\cite{liu2017dense} & - & - & - & 4.50 \\ 
		3DSTN~\cite{bhagavatula2017faster} & 3.15 & 4.33 & 5.98 & 4.49 \\ 
		MFN~\cite{chaudhuri2019joint} & 2.91 & 3.83 & 4.94 & 3.89 \\
		3DDFA-TPAMI~\cite{zhu2017face} & 2.84 & 3.57 & 4.96 & 3.79 \\ 
		PRNet~\cite{feng2018joint} & 2.75 & 3.51 & 4.61 & 3.62 \\
		RetinaFace~\cite{deng2020retinaface} & \textbf{2.57} & \textbf{3.32} & \textbf{4.56} & \textbf{3.48} \\  \hline
		Ours & 2.60 & 3.48 & 4.78 & 3.62 \\
		\hline 
	\end{tabular}
	}
	\label{tab_sparse_alignment}
\end{table}

\begin{figure}[htbp]
	\begin{center}
		\includegraphics[width=0.5\textwidth]{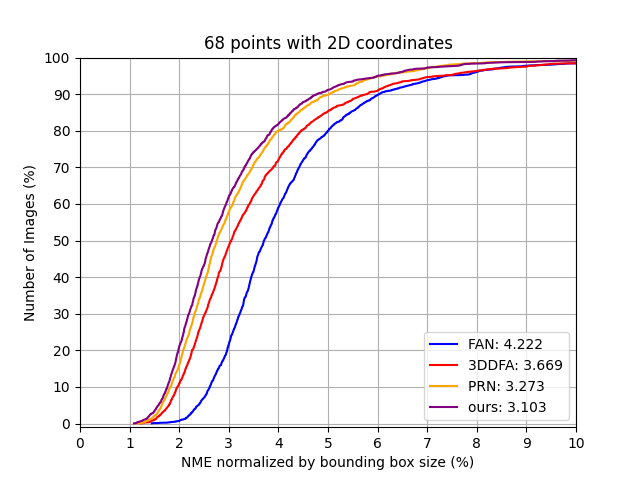}
	\end{center}
	\caption{Cumulative Errors Distribution (CED) curves on AFLW2000-3D. Evaluation is performed on 68 landmarks with 2D coordinates.}
	\label{fig: ced68}
\end{figure}

\textbf{Sparse Face Alignment.} Table~\ref{tab_sparse_alignment} shows the experimental results on small, medium, and large yaw angles. It can be seen that our proposed approach obtains a comparable result on facial landmark localization comparing to the state-of-the-art methods. Although it is not specifically designed for such task, our proposed method still outperforms the most similar approach~\cite{chaudhuri2019joint} jointly detecting and reconstructing 3D face shape.
Fig.~\ref{fig: ced68} plots the Cumulative Errors Distribution (CED) curves of $68$ landmarks on AFLW2000-3D. To facilitate fair comparisons, we use the officially released models~\cite{bulat2017far,feng2018joint,zhu2017face} on AFLW2000-3D dataset. It can be observed that our proposed approach performs better than the previous methods~\cite{bulat2017far,feng2018joint,zhu2017face}.

\begin{figure}[htbp]
	\begin{center}
		\includegraphics[width=0.5\textwidth]{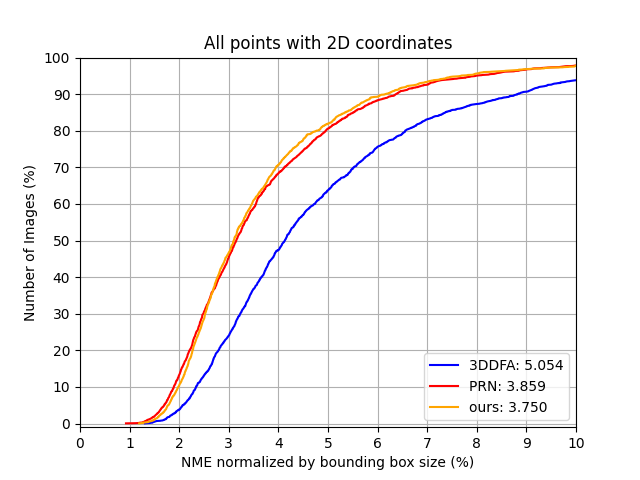}
	\end{center}
	\caption{Cumulative Errors Distribution (CED) curves on AFLW2000-3D. Evaluation is performed on all $2$D coordinates.}
	\label{fig: cedall}
\end{figure}

\textbf{Dense Face Alignment.} 
Fig.~\ref{fig: cedall} shows the CED curves of all $2$D points on AFLW2000-3D, in which the total of 35,709 vertices are evaluated. It can be clearly seen that our proposed achieves better results than those methods~\cite{feng2018joint,zhu2017face} taking advantage of dense 3D ground-truth data.

\begin{figure}[htbp]
	\begin{center}
		\includegraphics[width=0.5\textwidth]{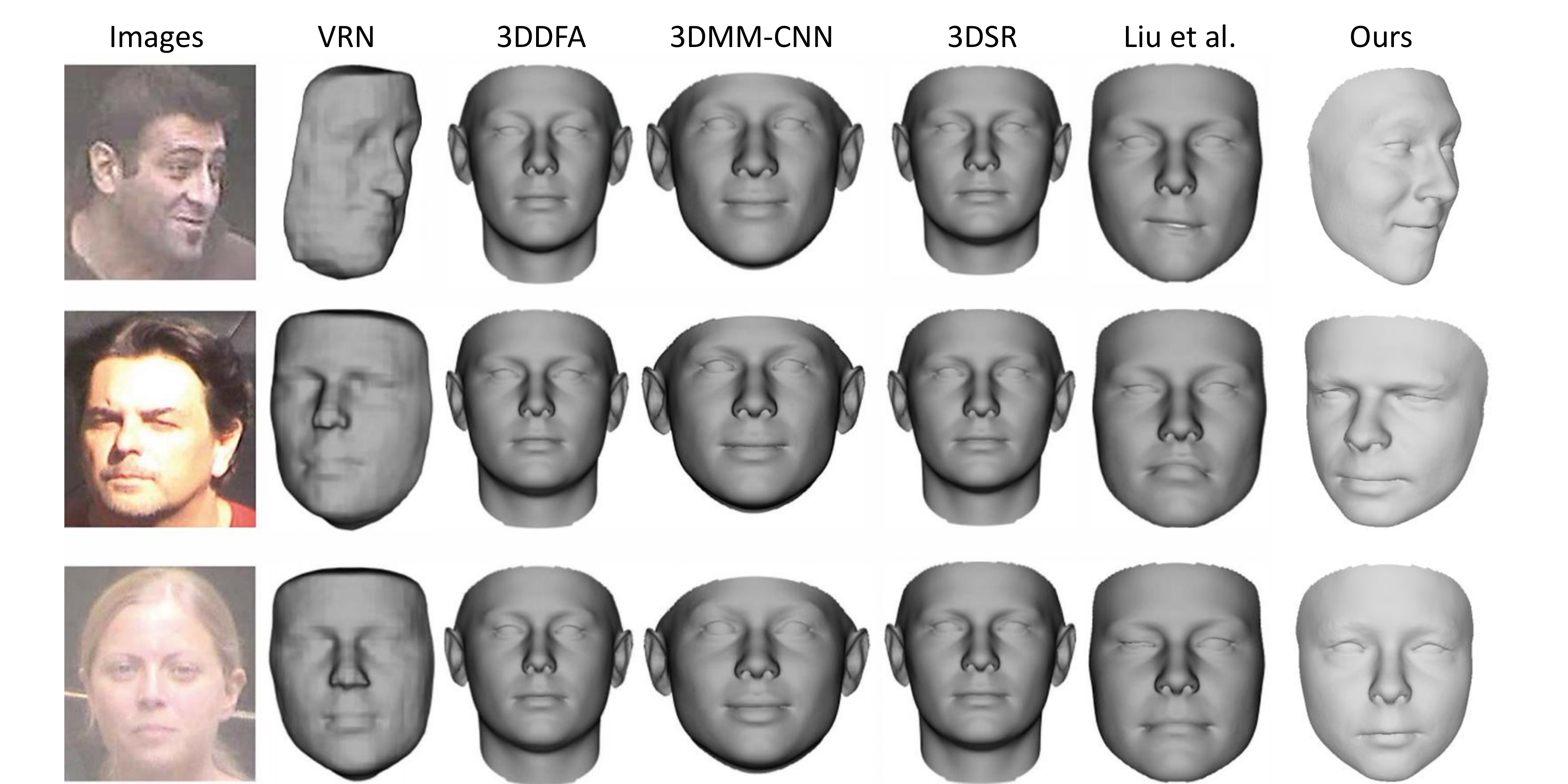}
	\end{center}
	\caption{Reconstruction results for three MICC subjects (first column). The rest columns show the reconstructed results of VRN\cite{jackson2017large}, $3$DDFA\cite{zhu2016face}, $3$DMM-CNN\cite{tran2017regressing}, $3$DSR\cite{liu2016joint}, Liu \emph{et~al.}\cite{liu2018disentangling} and the proposed method.}
	\label{fig:micc_example}
\end{figure}

\begin{figure}[htbp]
	\begin{center}
		\includegraphics[width=0.5\textwidth]{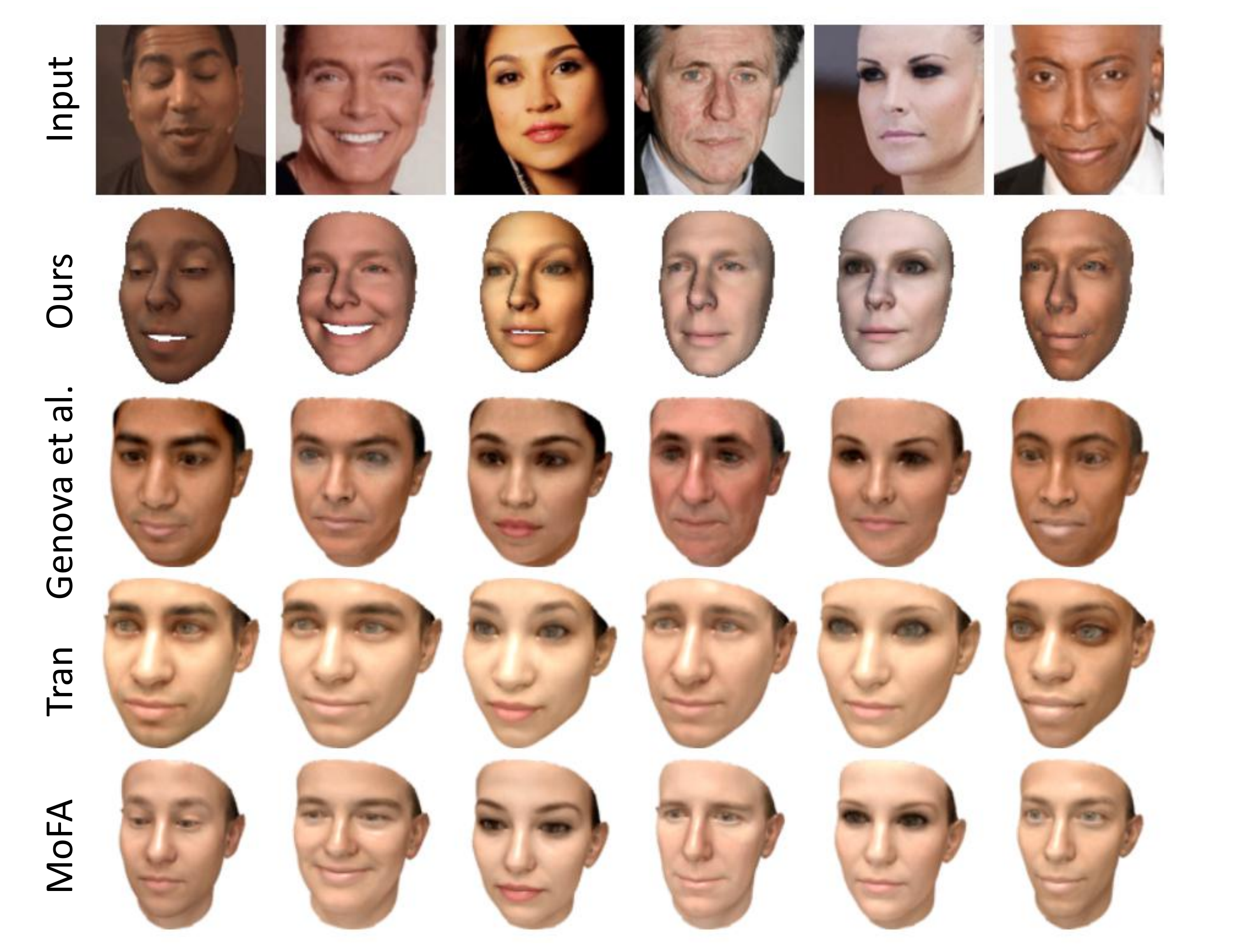}
	\end{center}
	\caption{Reconstruction results compared with MoFA\cite{tewari2017mofa}, Tran\cite{tran2017regressing} and Genova et al.\cite{genova2018unsupervised}. Our method shows improved likeness and color fidelity over competing methods.}
	\label{fig:qualitative}
\end{figure}

\begin{figure}[htbp]
	\begin{center}
		\includegraphics[width=1.\linewidth]{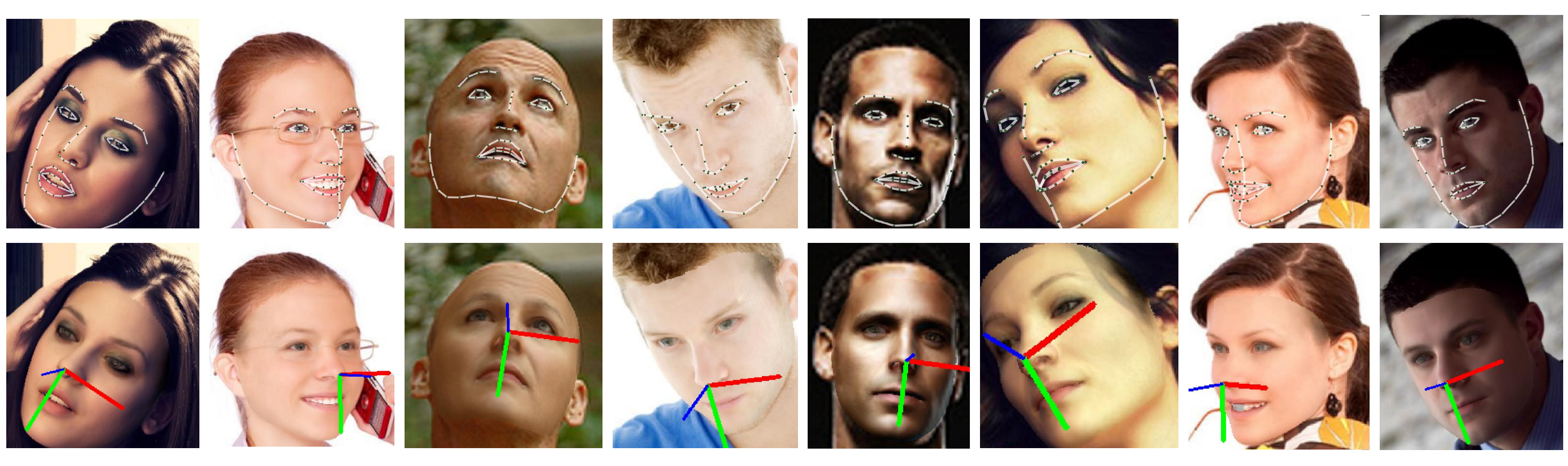}
	\end{center}
	\caption{Example results on AFLW2000-3D dataset. The top row shows the face alignment results, and the bottom is the reconstructed 3D faces as well as their poses.}
	\label{fig:face_alignment}
\end{figure}

\textbf{Qualitative Results.} Fig.~\ref{fig:micc_example} illustrates the qualitative results of the reconstructed faces compared against VRN~\cite{jackson2017large}, $3$DDFA~\cite{zhu2016face}, $3$DMM-CNN~\cite{tran2017regressing}, $3$DSR~\cite{liu2016joint} and Liu \emph{et~al.}~\cite{liu2018disentangling}. Moreover, we investigate the visual fidelity with MoFA~\cite{tewari2017mofa}, Tran~\cite{tran2017regressing} and Genova et al~\cite{genova2018unsupervised} in Fig.~\ref{fig:qualitative}. Additional visual example results are shown in Fig.~\ref{fig:face_alignment}, which indicate that our proposed framework is able to handle face reconstruction across large pose variations.
\begin{figure*}[htbp]
	\begin{center}
		\includegraphics[width=1\textwidth]{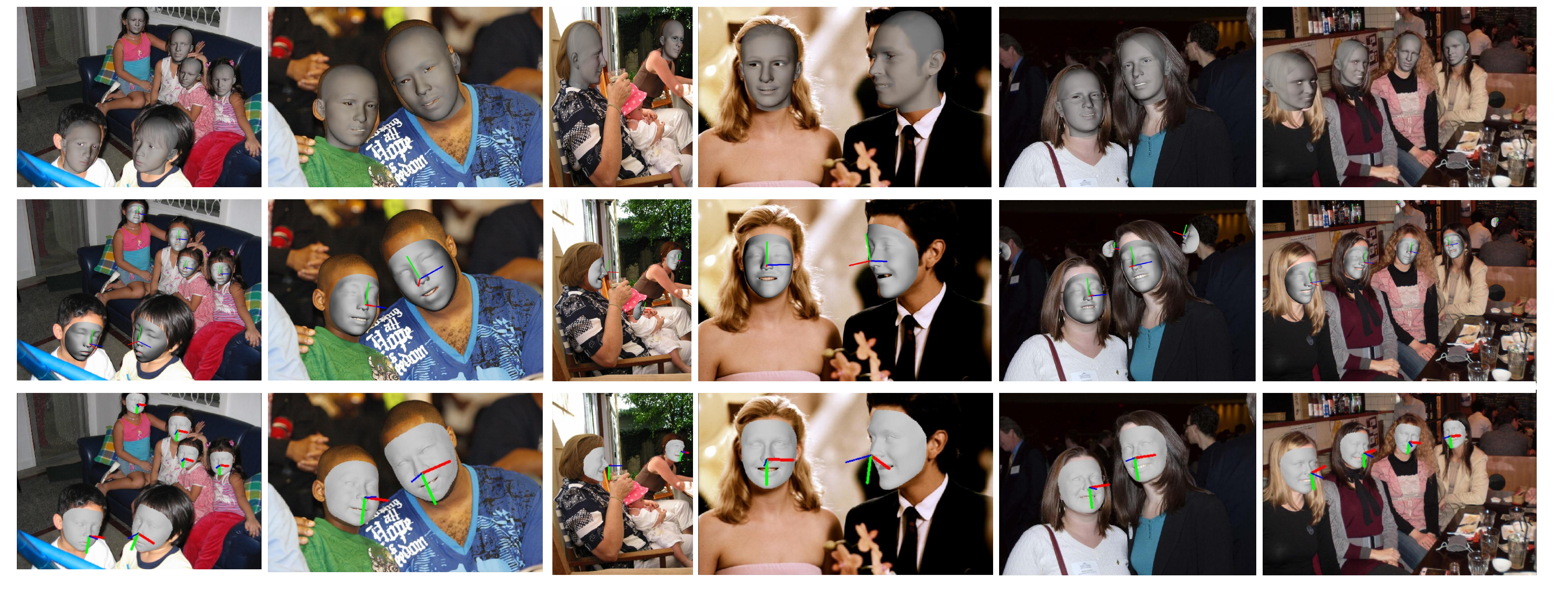}
	\end{center}
	\caption{Multi-face reconstruction, compared with Bindita~\cite{chaudhuri2019joint} (row 1) and Deng~\cite{deng2020retinaface} (row 2).}
	\label{fig:multi_shape}
\end{figure*}

\subsection{Results on Multi-Face Reconstruction}

\begin{figure}[t]
	\begin{center}
		\includegraphics[width=0.5\textwidth]{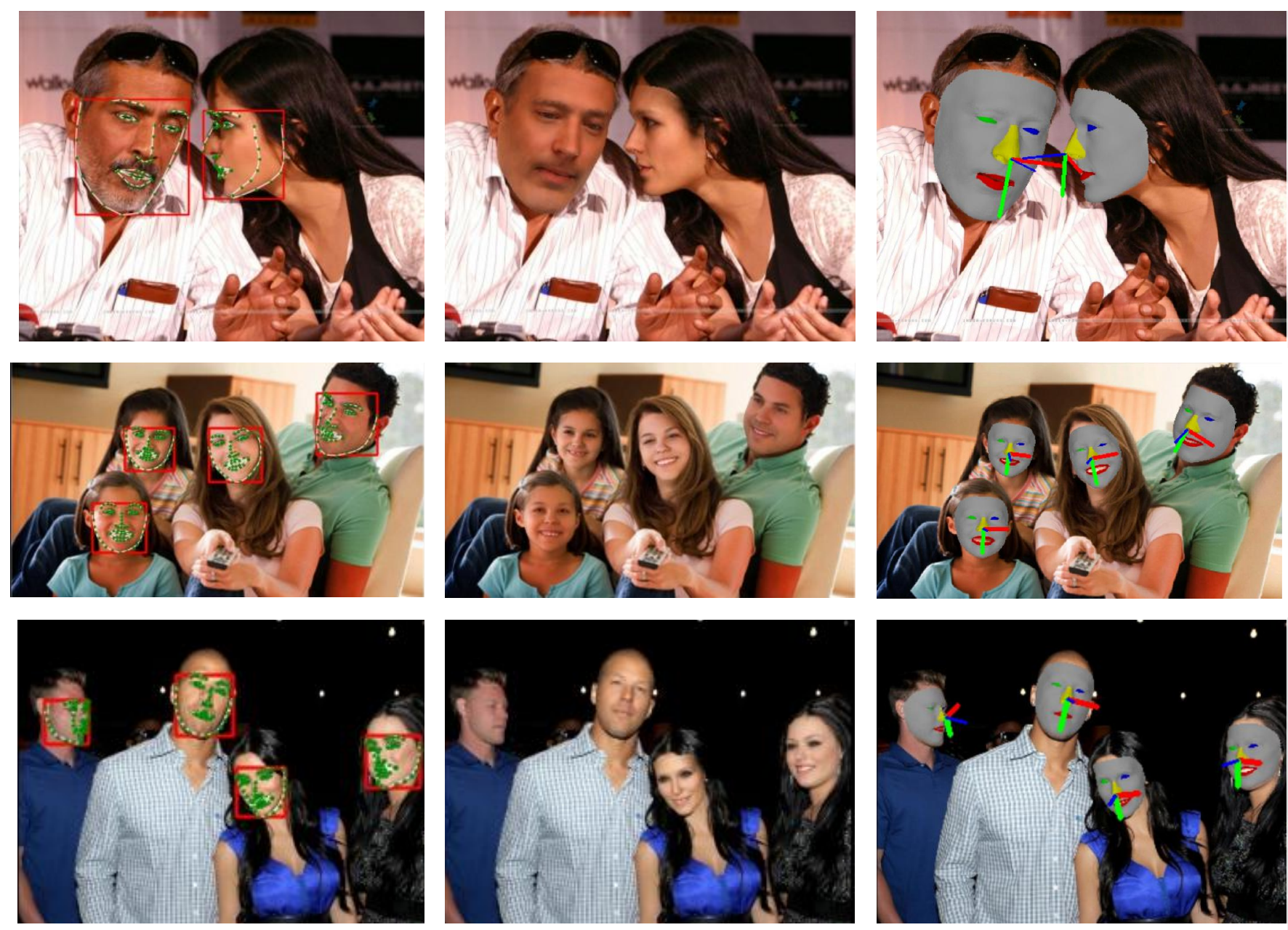}
	\end{center}
	\caption{Multi-face multi-level attributes. Column $1$: Input images with bounding boxes and landmarks. Column $2$: Rendered results. Column $3$: $3$D head poses and face segmentation results.}
	\label{fig:multi_level}
\end{figure}

\textbf{Multi-Face Reconstruction.} As shown in Fig.~\ref{fig:multi_shape}, we compare our proposed framework with the previous methods on multi-face images, which employ $3$D information for supervision.
Deng \emph{et al.}~\cite{deng2020retinaface} directly regress the dense face points on the image plane, which leads to better alignment result. Since the pose can not be estimated by the network, the least-squares estimation is applied for each face. Bindita \emph{et al.}~\cite{chaudhuri2019joint} predict the focal length for each face. Thus, the reconstructed $3$D faces are independent of each other, which loses the relative positions in camera space. Both of them focus on the shape reconstruction or vertex regression while our framework takes the albedo and lighting into modeling further. As a consequence, we can reconstruct all faces with texture as shown in the second column of Fig.~\ref{fig:multi_level} with high similarity and fidelity. 

\begin{figure}[htbp]
	\begin{center}
		\includegraphics[width=0.5\textwidth]{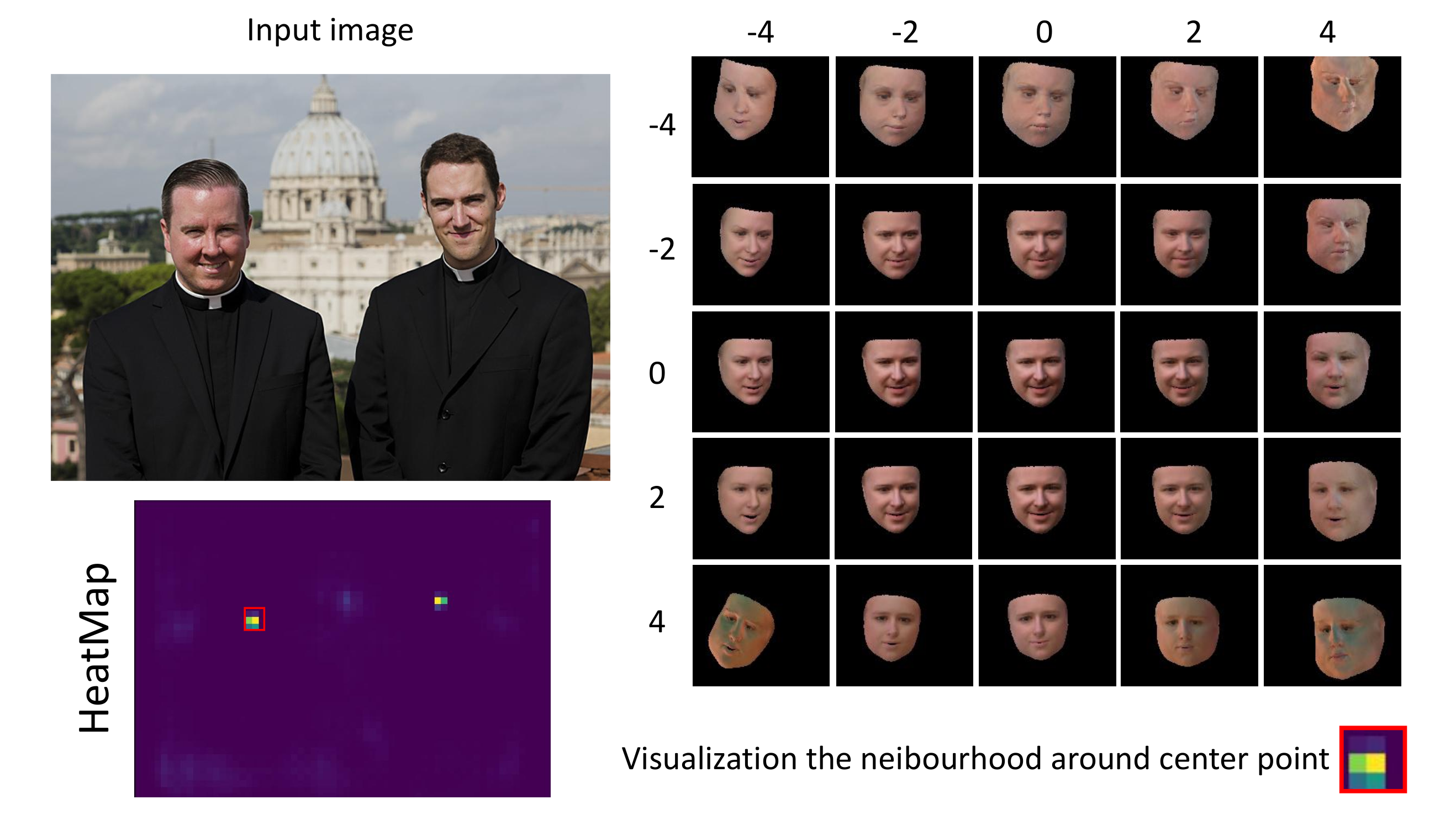}
	\end{center}
	\caption{Visualization of the reconstructed faces around the center.}
	\label{fig:feature}
\end{figure}

\textbf{Feature Map.} It is also interesting to analyze the regressed face feature maps. As shown in Fig.~\ref{fig:feature}, the feature points near the center have similar reconstructed faces. It makes our proposed framework more robust comparing to the conventional pipeline whose reconstructed quality is sensitive to their pre-processing step.

\begin{figure}[t]
	\begin{center}
		\includegraphics[width=.45\textwidth]{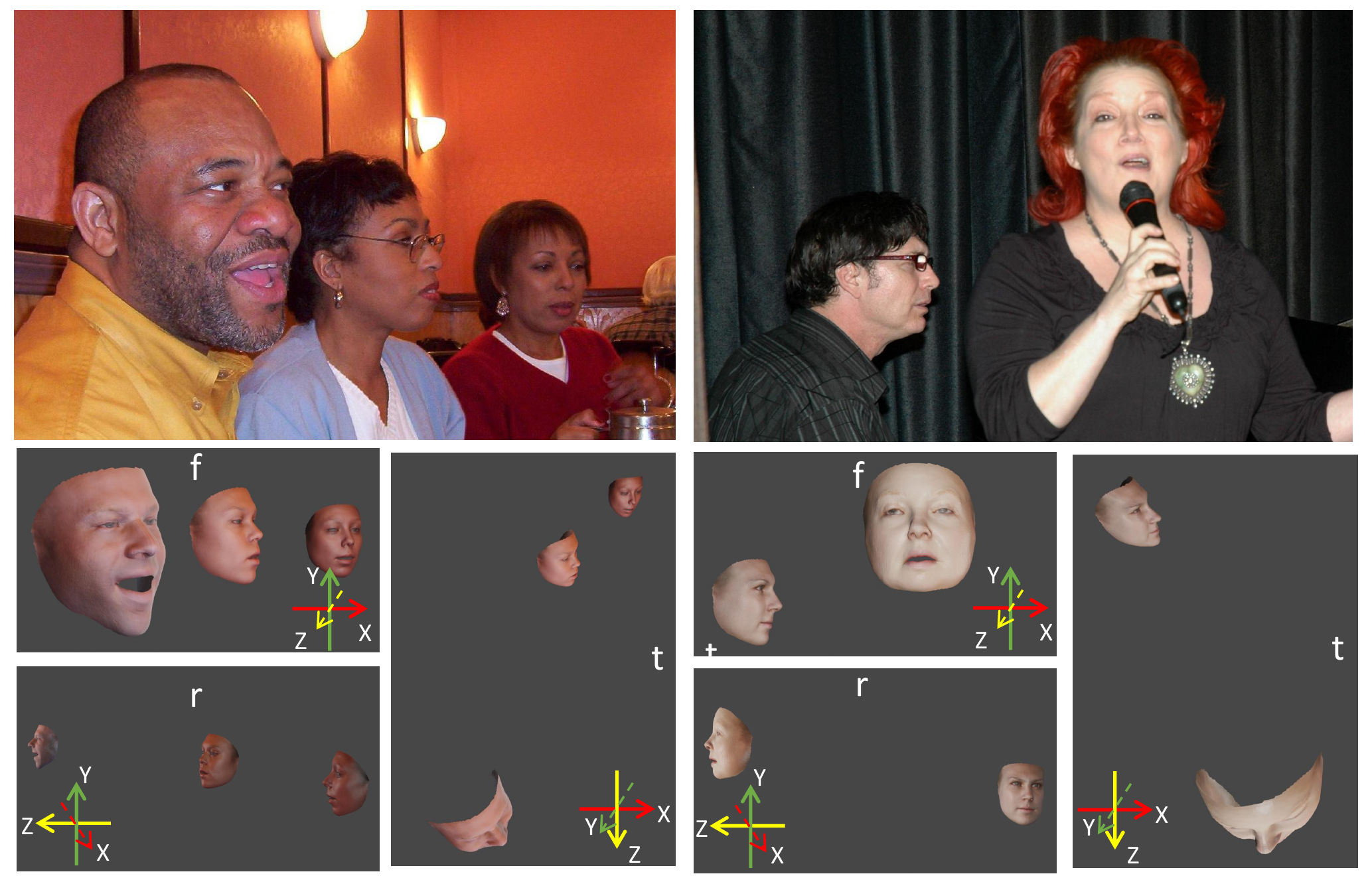}
	\end{center}
	\caption{Global view of reconstructed multiple faces. f: front view, r: right view and t: top view are shown in the figure.}
	\label{fig:projection}
\end{figure}

\textbf{Global Camera Model.} Fig.~\ref{fig:projection} demonstrates some examples of the reconstructed faces from different views. As we employ a global perspective projection model for all faces, the reconstructed faces in the image are under the same camera view. It can be seen that their relative positions are retained. 

\textbf{Facial Attributes.} With the recovered 3D face models, our proposed approach is able to predict the bounding boxes, facial landmarks, head poses, and face components segmentation, as shown in Fig.~\ref{fig:multi_level}.

\subsection{Comparison on Efficiency} 
Finally, we discuss the efficiency of our proposed framework. The computational time is measured on a PC with $2080$Ti GPU, E$5-2678$ CPU @ $2.50$GHz. Our method is implemented in PyTorch.

In our experiments, it only costs $12$ms for our proposed approach to predict the face locations and their parameters with the input size of $512\times512$. For the conventional pipeline~\cite{deng2019accurate} using the same ResNet-50 backbone, it takes $8$ms to encode a single face. When the number of faces in an image increases to $5$ and $10$, the computational time for parameter estimation is around $40$ms and $80$ms, respectively. Note that the pre-processing time on cropping and resizing becomes non-negligible for the conventional pipeline to deal with multiple faces. It can be clearly observed that our proposed single shot multi-face reconstruction framework enjoys the merit of constant computational time on predicting the 3D model parameters with the increased number of faces in image. This is because our proposed approach significantly simplifies the whole 3D face recovery pipeline comparing to the conventional methods by sharing the computation on deep learning feature extraction in localization and model estimation. More importantly, our method only has single neural network model, which makes it much easier to deploy in real-world applications. 

%% file: conclusion.tex
\section{Conclusions}

In this paper, we have proposed the effective single-shot end-to-end framework for multi-face reconstruction, which is able to predict the model parameters for multiple instances using single network inference. Compared to the conventional reconstruction pipeline, our proposed approach not only greatly reduces the computational redundancy on feature extraction but also simplifies the deployment procedure. By employing a global camera model, the relative $3$D positions and orientations of reconstructed faces can be recovered. The experimental results indicate that our proposed approach is very promising on face alignment tasks without fully-supervision and pre-processing like detection and crop. For the future work, we will extend our proposed framework for the generic object reconstruction by exploring the effective deformation models.

%% file: main.bbl
\begin{thebibliography}{10}

\bibitem{bas20173d}
A.~Bas, P.~Huber, W.~A. Smith, M.~Awais, and J.~Kittler.
\newblock 3d morphable models as spatial transformer networks.
\newblock In {\em International Conference on Computer Vision Workshop on
  Geometry Meets Deep Learning}, pages 904--912, 2017.

\bibitem{bhagavatula2017faster}
C.~Bhagavatula, C.~Zhu, K.~Luu, and M.~Savvides.
\newblock Faster than real-time facial alignment: A 3d spatial transformer
  network approach in unconstrained poses.
\newblock In {\em Proceedings of the IEEE International Conference on Computer
  Vision}, pages 3980--3989, 2017.

\bibitem{blanz1999morphable}
V.~Blanz and T.~Vetter.
\newblock A morphable model for the synthesis of 3d faces.
\newblock In {\em Proceedings of the 26th annual conference on Computer
  graphics and interactive techniques}, pages 187--194, 1999.

\bibitem{bulat2017far}
A.~Bulat and G.~Tzimiropoulos.
\newblock How far are we from solving the 2d \& 3d face alignment problem?(and
  a dataset of 230,000 3d facial landmarks).
\newblock In {\em Proceedings of the IEEE International Conference on Computer
  Vision}, pages 1021--1030, 2017.

\bibitem{cao2013facewarehouse}
C.~Cao, Y.~Weng, S.~Zhou, Y.~Tong, and K.~Zhou.
\newblock Facewarehouse: A 3d facial expression database for visual computing.
\newblock {\em IEEE Transactions on Visualization and Computer Graphics},
  20(3):413--425, 2013.

\bibitem{cao2014face}
X.~Cao, Y.~Wei, F.~Wen, and J.~Sun.
\newblock Face alignment by explicit shape regression.
\newblock {\em International Journal of Computer Vision}, 107(2):177--190,
  2014.

\bibitem{chaudhuri2019joint}
B.~Chaudhuri, N.~Vesdapunt, and B.~Wang.
\newblock Joint face detection and facial motion retargeting for multiple
  faces.
\newblock In {\em Proceedings of the IEEE Conference on Computer Vision and
  Pattern Recognition}, pages 9719--9728, 2019.

\bibitem{deng2020retinaface}
J.~Deng, J.~Guo, E.~Ververas, I.~Kotsia, and S.~Zafeiriou.
\newblock Retinaface: Single-shot multi-level face localisation in the wild.
\newblock In {\em Proceedings of the IEEE/CVF Conference on Computer Vision and
  Pattern Recognition}, pages 5203--5212, 2020.

\bibitem{deng2019accurate}
Y.~Deng, J.~Yang, S.~Xu, D.~Chen, Y.~Jia, and X.~Tong.
\newblock Accurate 3d face reconstruction with weakly-supervised learning: From
  single image to image set.
\newblock In {\em Proceedings of the IEEE Conference on Computer Vision and
  Pattern Recognition Workshops}, pages 0--0, 2019.

\bibitem{dou2017end}
P.~Dou, S.~K. Shah, and I.~A. Kakadiaris.
\newblock End-to-end 3d face reconstruction with deep neural networks.
\newblock In {\em IEEE Conference on Computer Vision and Pattern Recognition
  (CVPR)}, pages 21--26, 2017.

\bibitem{feng2018joint}
Y.~Feng, F.~Wu, X.~Shao, Y.~Wang, and X.~Zhou.
\newblock Joint 3d face reconstruction and dense alignment with position map
  regression network.
\newblock In {\em Proceedings of the European Conference on Computer Vision
  (ECCV)}, pages 534--551, 2018.

\bibitem{fyffe2014driving}
G.~Fyffe, A.~Jones, O.~Alexander, R.~Ichikari, and P.~Debevec.
\newblock Driving high-resolution facial scans with video performance capture.
\newblock {\em ACM Transactions on Graphics (TOG)}, 34(1):1--14, 2014.

\bibitem{genova2018unsupervised}
K.~Genova, F.~Cole, A.~Maschinot, A.~Sarna, D.~Vlasic, and W.~T. Freeman.
\newblock Unsupervised training for 3d morphable model regression.
\newblock In {\em IEEE Conference on Computer Vision and Pattern Recognition
  (CVPR)}, June 2018.

\bibitem{Genova_2018_CVPR}
K.~Genova, F.~Cole, A.~Maschinot, A.~Sarna, D.~Vlasic, and W.~T. Freeman.
\newblock Unsupervised training for 3d morphable model regression.
\newblock In {\em The IEEE Conference on Computer Vision and Pattern
  Recognition (CVPR)}, June 2018.

\bibitem{he2016deep}
K.~He, X.~Zhang, S.~Ren, and J.~Sun.
\newblock Deep residual learning for image recognition.
\newblock In {\em Proceedings of the IEEE conference on computer vision and
  pattern recognition}, pages 770--778, 2016.

\bibitem{henderson19ijcv}
P.~Henderson and V.~Ferrari.
\newblock Learning single-image {3D} reconstruction by generative modelling of
  shape, pose and shading.
\newblock {\em International Journal of Computer Vision}, 2019.

\bibitem{jackson2017large}
A.~S. Jackson, A.~Bulat, V.~Argyriou, and G.~Tzimiropoulos.
\newblock Large pose 3d face reconstruction from a single image via direct
  volumetric cnn regression.
\newblock In {\em Proceedings of the IEEE International Conference on Computer
  Vision}, pages 1031--1039, 2017.

\bibitem{kingma2014adam}
D.~P. Kingma and J.~Ba.
\newblock Adam: A method for stochastic optimization.
\newblock {\em arXiv preprint arXiv:1412.6980}, 2014.

\bibitem{liu2016joint}
F.~Liu, D.~Zeng, Q.~Zhao, and X.~Liu.
\newblock Joint face alignment and 3d face reconstruction.
\newblock In {\em European Conference on Computer Vision}, pages 545--560.
  Springer, 2016.

\bibitem{liu2018disentangling}
F.~Liu, R.~Zhu, D.~Zeng, Q.~Zhao, and X.~Liu.
\newblock Disentangling features in 3d face shapes for joint face
  reconstruction and recognition.
\newblock In {\em Proceedings of the IEEE conference on computer vision and
  pattern recognition}, pages 5216--5225, 2018.

\bibitem{liu2017dense}
Y.~Liu, A.~Jourabloo, W.~Ren, and X.~Liu.
\newblock Dense face alignment.
\newblock In {\em Proceedings of the IEEE International Conference on Computer
  Vision Workshops}, pages 1619--1628, 2017.

\bibitem{liu2015deep}
Z.~Liu, P.~Luo, X.~Wang, and X.~Tang.
\newblock Deep learning face attributes in the wild.
\newblock In {\em Proceedings of the IEEE international conference on computer
  vision}, pages 3730--3738, 2015.

\bibitem{paysan20093d}
P.~Paysan, R.~Knothe, B.~Amberg, S.~Romdhani, and T.~Vetter.
\newblock A 3d face model for pose and illumination invariant face recognition.
\newblock In {\em 2009 Sixth IEEE International Conference on Advanced Video
  and Signal Based Surveillance}, pages 296--301. Ieee, 2009.

\bibitem{ravi2020pytorch3d}
N.~Ravi, J.~Reizenstein, D.~Novotny, T.~Gordon, W.-Y. Lo, J.~Johnson, and
  G.~Gkioxari.
\newblock Accelerating 3d deep learning with pytorch3d.
\newblock {\em arXiv:2007.08501}, 2020.

\bibitem{richardson20163d}
E.~Richardson, M.~Sela, and R.~Kimmel.
\newblock 3d face reconstruction by learning from synthetic data.
\newblock In {\em International Conference on 3D Vision (3DV)}, pages 460--469,
  2016.

\bibitem{schroff2015facenet}
F.~Schroff, D.~Kalenichenko, and J.~Philbin.
\newblock Facenet: A unified embedding for face recognition and clustering.
\newblock In {\em Proceedings of the IEEE conference on computer vision and
  pattern recognition}, pages 815--823, 2015.

\bibitem{sengupta2018sfsnet}
S.~Sengupta, A.~Kanazawa, C.~D. Castillo, and D.~W. Jacobs.
\newblock Sfsnet: Learning shape, reflectance and illuminance of facesin the
  wild'.
\newblock In {\em Proceedings of the IEEE Conference on Computer Vision and
  Pattern Recognition}, pages 6296--6305, 2018.

\bibitem{shen2015first}
J.~Shen, S.~Zafeiriou, G.~G. Chrysos, J.~Kossaifi, G.~Tzimiropoulos, and
  M.~Pantic.
\newblock The first facial landmark tracking in-the-wild challenge: Benchmark
  and results.
\newblock In {\em Proceedings of the IEEE international conference on computer
  vision workshops}, pages 50--58, 2015.

\bibitem{tewari2018self}
A.~Tewari, M.~Zollh{\"o}fer, P.~Garrido, F.~Bernard, H.~Kim, P.~P{\'e}rez, and
  C.~Theobalt.
\newblock Self-supervised multi-level face model learning for monocular
  reconstruction at over 250 hz.
\newblock In {\em IEEE Conference on Computer Vision and Pattern Recognition
  (CVPR)}, pages 2549--2559, 2018.

\bibitem{tewari2017mofa}
A.~Tewari, M.~Zollh{\"o}fer, H.~Kim, P.~Garrido, F.~Bernard, P.~P{\'e}rez, and
  C.~Theobalt.
\newblock {MoFa:} model-based deep convolutional face autoencoder for
  unsupervised monocular reconstruction.
\newblock In {\em International Conference on Computer Vision (ICCV)}, pages
  1274--1283, 2017.

\bibitem{thies2016face2face}
J.~Thies, M.~Zollhofer, M.~Stamminger, C.~Theobalt, and M.~Nie{\ss}ner.
\newblock Face2face: Real-time face capture and reenactment of rgb videos.
\newblock In {\em Proceedings of the IEEE conference on computer vision and
  pattern recognition}, pages 2387--2395, 2016.

\bibitem{tran2017regressing}
A.~T. Tran, T.~Hassner, I.~Masi, and G.~Medioni.
\newblock Regressing robust and discriminative 3d morphable models with a very
  deep neural network.
\newblock In {\em IEEE Conference on Computer Vision and Pattern Recognition
  (CVPR)}, pages 1493--1502, 2017.

\bibitem{tran2018nonlinear}
L.~Tran and X.~Liu.
\newblock Nonlinear 3d face morphable model.
\newblock {\em IEEE Conference on Computer Vision and Pattern Recognition
  (CVPR)}, pages 7346--7355, 2018.

\bibitem{wu2020unsupervised}
S.~Wu, C.~Rupprecht, and A.~Vedaldi.
\newblock Unsupervised learning of probably symmetric deformable 3d objects
  from images in the wild.
\newblock In {\em Proceedings of the IEEE/CVF Conference on Computer Vision and
  Pattern Recognition}, pages 1--10, 2020.

\bibitem{xiang2017joint}
J.~Xiang and G.~Zhu.
\newblock Joint face detection and facial expression recognition with mtcnn.
\newblock In {\em 2017 4th International Conference on Information Science and
  Control Engineering (ICISCE)}, pages 424--427. IEEE, 2017.

\bibitem{xiong2015global}
X.~Xiong and F.~De~la Torre.
\newblock Global supervised descent method.
\newblock In {\em Proceedings of the IEEE Conference on Computer Vision and
  Pattern Recognition}, pages 2664--2673, 2015.

\bibitem{yang2016wider}
S.~Yang, P.~Luo, C.-C. Loy, and X.~Tang.
\newblock Wider face: A face detection benchmark.
\newblock In {\em Proceedings of the IEEE conference on computer vision and
  pattern recognition}, pages 5525--5533, 2016.

\bibitem{yu2017learning}
R.~Yu, S.~Saito, H.~Li, D.~Ceylan, and H.~Li.
\newblock Learning dense facial correspondences in unconstrained images.
\newblock In {\em Proceedings of the IEEE International Conference on Computer
  Vision}, pages 4723--4732, 2017.

\bibitem{zhou2019talking}
H.~Zhou, Y.~Liu, Z.~Liu, P.~Luo, and X.~Wang.
\newblock Talking face generation by adversarially disentangled audio-visual
  representation.
\newblock In {\em Proceedings of the AAAI Conference on Artificial
  Intelligence}, volume~33, pages 9299--9306, 2019.

\bibitem{zhou2019objects}
X.~Zhou, D.~Wang, and P.~Kr{\"a}henb{\"u}hl.
\newblock Objects as points.
\newblock {\em arXiv preprint arXiv:1904.07850}, 2019.

\bibitem{zhu2016face}
X.~Zhu, Z.~Lei, X.~Liu, H.~Shi, and S.~Z. Li.
\newblock Face alignment across large poses: A 3d solution.
\newblock In {\em Proceedings of the IEEE conference on computer vision and
  pattern recognition}, pages 146--155, 2016.

\bibitem{zhu2017face}
X.~Zhu, X.~Liu, Z.~Lei, and S.~Z. Li.
\newblock Face alignment in full pose range: A 3d total solution.
\newblock {\em IEEE transactions on pattern analysis and machine intelligence},
  41(1):78--92, 2017.

\bibitem{zulqarnain2018learning}
S.~Zulqarnain~Gilani and A.~Mian.
\newblock Learning from millions of 3d scans for large-scale 3d face
  recognition.
\newblock In {\em Proceedings of the IEEE Conference on Computer Vision and
  Pattern Recognition}, pages 1896--1905, 2018.

\end{thebibliography}
